\documentclass[letterpaper, 10 pt, conference]{ieeeconf} 

\usepackage[english]{babel}											%
\usepackage[T1]{fontenc}											%
\usepackage{textcomp}												%
\usepackage{dsfont}													%
\usepackage{booktabs}												%
\usepackage{array}													%
\usepackage{bm}														%
\usepackage{url}													%
\usepackage{enumerate}												%
\usepackage[table]{xcolor}											%
\usepackage{amssymb, amsfonts}								%
\usepackage{amsmath}
\usepackage{cases}													%
\usepackage{ifthen}													%
\usepackage{ifpdf}													%
\usepackage[pdftex]{graphicx}										%
\DeclareGraphicsExtensions{.pdf,.png,.jpg,.mps}						%
\usepackage{framed}													%
\usepackage[amsmath,thmmarks,framed,hyperref]{ntheorem}				%
\usepackage{eso-pic}												%
\usepackage[switch, pagewise, mathlines]{lineno}					%
\usepackage[color=white!90!black, textwidth=3cm]{todonotes}			%
\usepackage{algorithm}												%
\usepackage[noend]{algpseudocode}									%
\usepackage{acronym}												%
\usepackage{tikz}				
\usetikzlibrary{calc}												%
\usetikzlibrary{arrows,decorations.text,decorations.pathmorphing}	%
\usetikzlibrary{shadows,patterns,positioning,shapes,matrix,fit}		%
\usetikzlibrary{pgfplots.groupplots}
\usepackage{gensymb}
\usepackage{pgfplots}
\usepackage{adjustbox}
\usepackage{color,soul} 

\usepgfplotslibrary{fillbetween}
\usepackage{graphicx}
							
\pdfoutput=1
\newcounter{generalCounter}
\theoremclass		{Theorem}
\theoremstyle		{plain}
\theoremheaderfont	{\normalfont\bfseries}
\theorembodyfont	{\normalfont}
\theorempostwork	{\hrule}

\theoremstyle		{nonumberplain}
\theorempostwork	{\hrule}
\theoremseparator	{}
\theoremheaderfont	{\normalfont\bfseries}
\theorembodyfont	{\normalfont}
\theoremsymbol		{\ensuremath{\diamondsuit}}

\graphicspath{{../Images/}}

\pgfdeclarelayer{ultrabackground}
\pgfdeclarelayer{background}
\pgfdeclarelayer{foreground}
\pgfsetlayers{ultrabackground,background,main,foreground}

\def\TablesColumnsColor{black!4}

\newcolumntype
{g}
{
	>{\centering \columncolor{\TablesColumnsColor} \arraybackslash}
	p{0.15\textwidth}
	<{}
}

\newcolumntype
{w}
{
	>{\centering \arraybackslash}
	p{0.15\textwidth}
	<{}
}

\newcommand{\GammaDistribution}				[2]	{\Gamma \left( #1, #2 \right)}

\newcommand{\DefinedAs}			[0]	{\mathrel{\mathop:}=}

\newcommand{\ProbabilityDensity}			[0]	{p}
\newcommand{\ProbabilityDensityOf}			[1]	{\ProbabilityDensity \left( #1 \right)}
\newcommand{\ProbabilityDensityOfGiven}		[2]	{\ProbabilityDensityOf{ #1 \left| #2 \right. }}

\newcommand{\Expectation}					[0]	{\mathbb{E}}
\newcommand{\ExpectationOf}					[1]	{\Expectation \left[ #1 \right]}

\newcommand{\ExpectationOfGiven}			[2]	{\ExpectationOf{ #1 \; \left| \; #2 \right. }}

\newcommand	{\Section}				[0]	{Sec.}
\newcommand	{\Sections}				[0]	{Sections}
\newcommand	{\Equation}				[0]	{Equ.}

\newcommand	{\Figure}				[0]	{Fig.}

\newcommand	{\Table}				[0]	{Tab.}

\acrodef{RMSE}		[RMSE]			{Root-Mean-Square Error}

\acrodef{amcw}		[AMCW]			{Amplitude Modulated Continuous Wave}
\acrodef{AIC}		[AIC]			{Akaike Information Criterion}
\acrodef{BLUE}		[BLUE]			{Best Linear Unbiased Estimator}
\acrodef{CDF}		[CDF]			{Cumulative Distribution Function}
\acrodef{ce}		[CE]			{Cross Entropy}
\acrodef{ci}		[CI]			{Covariance Intersection}
\acrodef{CR}		[C-R]			{Cram\'er-Rao}
\acrodef{cu}		[CU]			{Covariance Union}
\acrodef{cw}		[CW]			{Continuous Wave}
\acrodef{EM}		[EM]			{Expectation Maximization}
\acrodef{fmcw}		[FMCW]			{Frequency Modulated Continuous Wave}
\acrodef{gcc}		[GCC]			{Generalized Convex Combination}
\acrodef{GP}		[GP]			{Gaussian Process}
\acrodef{GLR}		[GLR]			{Generalized Likelihood Ratio}
\acrodef{gmm}		[GMM]			{Gaussian mixture model}
\acrodef{iid}		[iid]			{independent and identically distributed}
\acrodef{LMMSE}		[LMMSE]			{Linear Minimum Mean Square Error}
\acrodef{LS}		[LS]			{Least Squares}
\acrodef{lidar}		[Lidar]			{Light Detection and Ranging}
\acrodef{lti}		[LTI]			{Linear Time Invariant Systems}
\acrodef{MAP}		[MAP]			{Maximum A Posteriori}
\acrodef{MCMC}		[MCMC]			{Markov Chain Monte Carlo}
\acrodef{ML}		[ML]			{Maximum Likelihood}
\acrodef{MMSE}		[MMSE]			{Minimum Mean Square Error}
\acrodef{MSE}		[MSE]			{Mean Squared Error}
\acrodef{MVUE}		[MVUE]			{Minimum Variance Unbiased Estimator}
\acrodef{NARX}		[NARX]			{Nonlinear AutoRegressive eXogenus}
\acrodef{NCS}		[NCS]			{Networked Control System}
\acrodef{NDT}		[NDT]			{Normal Distribution Transform}
\acrodef{nrmse}		[NRMSE]			{Normalized Root Mean Square Error}
\acrodef{PDF}		[PDF]			{Probability Density Function}
\acrodef{pem}		[PEM]			{Prediction Error Method}
\acrodef{RKHS}		[RKHS]			{Reproducing Kernel Hilbert Space}
\acrodef{RMSE}		[RMSE]			{Root-Mean-Square Error}
\acrodef{RN}		[RN]			{Regularization Network}
\acrodef{slam}		[SLAM]			{Simultaneous localization and mapping}
\acrodef{rvv}		[r.\textbf{v}.]	{random vector}
\acrodef{rv}		[r.v.]			{random variable}
\acrodef{tof}		[ToF]			{Time of Flight}
\acrodef{vicon}		[MoCap]			{Motion Capture}
\acrodef{WSN}		[WSN]			{Wireless Sensor Network}
\acrodef{MCMC}		[MCMC]			{Markov chain Monte Carlo}
\acrodef{FMCW}		[FMCW]			{Frequency-Modulated Continuous Wave}
\acrodef{FMCWr}		[FMCW radar]	{Frequency-Modulated Continuous Wave radar}
\acrodef{MPR}		[MPR]			{Mechanically Pivoting Radar}
\acrodef{FFT}		[FFT]			{Fast Fourier Transform}
\acrodef{IF}		[IF]			{Intermediate Frequency}
\acrodef{CFAR}		[CFAR]			{Constant False-Alarm Rate}
\acrodef{CACFAR}	[CA-CFAR]		{Cell-averaging CFAR}
\acrodef{ICP}		[ICP]			{Iterative closest point}
\acrodef{SINR}		[SINR]			{signal-to-interference-plus-noise ratio}
\acrodef{SNR}		[SNR]			{signal-to-noise ratio}
\acrodef{CFAR}		[CFAR]			{Constant False-Alarm Rate}
\acrodef{BFAR}		[BFAR]			{Bounded False-Alarm Rate}
\acrodef{FAR}		[FAR]			{False-Alarm Rate}
\acrodef{CACFAR}	[CA-CFAR]		{Cell-Averaging CFAR}
\acrodef{SOCACFAR}	[SOCA-CFAR]		{Smallest Of Cell Averaging CFAR}
\acrodef{GOCACFAR}	[GOCA-CFAR]		{Greatest Of Cell Averaging CFAR}
\acrodef{TMCFAR}	[TM-CFAR]		{Trimmed Mean CFAR}
\acrodef{CSCFAR}	[CS-CFAR]		{Censored CFAR}
\acrodef{OSCFAR}	[OS-CFAR]		{Ordered Statistics CFAR}
\acrodef{ACMLCFAR}	[ACML-CFAR]		{Adaptive Censoring Maximum Likelihood CFAR}
\acrodef{MLCFAR}	[ML-CFAR]		{Maximum Likelihood CFAR}
\acrodef{CMLDCFAR}	[CMLD-CFAR]		{Censored Mean Level Detector CFAR}
\acrodef{VICFAR}	[VI-CFAR]		{Variability Index CFAR}
\acrodef{ACCACFAR}	[ACCA-CFAR]		{Automatic Censored Cell Averagin CFAR}
\acrodef{CUT}	[CUT]		{Cell Under Test}
\acrodef{PD}	[$P_D$]		{Probability of Detection}
\acrodef{PFA}	[$P_\mathit{FA}$]		{Probability of False-Alarm}
\acrodef{PFAB}	[$P_\mathit{FAub}$]	{Probability of False-Alarm upper-bound}
\acrodef{ATE}	[ATE]	{Absolute Trajectory Error}
\acrodef{RPE}	[RPE]	{Relative Pose Error}
\acrodef{CFEAR}	[CFEAR]	{Conservative Filtering for Efficient and Accurate Radar odometry}
\acrodef{GPS}	[GPS]	{Global Positioning System}
\acrodef{MGF}	[mgf]	{moment generating function}

\definecolor{bblue}{HTML}{608dc3}
\definecolor{rred}{HTML}{c6615e}
\definecolor{ggreen}{HTML}{a5c169}
\definecolor{ppurple}{HTML}{bb81a3}
\definecolor{yyelow}{HTML}{FDAE61}

\definecolor{clr1}{RGB}{0,69,134}
\definecolor{clr2}{RGB}{255,66,14}
\definecolor{clr3}{RGB}{255,211,32}
\definecolor{clr4}{RGB}{87,157,28}
\definecolor{clr5}{RGB}{126,0,33}
\definecolor{clr6}{RGB}{131,202,255}
\definecolor{clr7}{RGB}{49,64,4}

\IEEEoverridecommandlockouts 

\usepackage{cite} 
\usepackage{todonotes}

\title{\LARGE \bf
BFAR -- Bounded False Alarm Rate detector for improved radar odometry estimation
}

\author{Anas Alhashimi,$^{1,2}$ Daniel Adolfsson,$^{1}$ Martin Magnusson,$^{1}$ Henrik Andreasson,$^{1}$ and Achim J. Lilienthal$^{1}$
\thanks{$^{1}$Center for Applied Autonomous Sensor Systems (AASS), {\"O}rebro University, {\"O}rebro, Sweden
 {\tt\small \{anas.alhashimi,martin.magnusson\}@oru.se}}%
\thanks{$^{2}$Computer Engineering Department, University of Baghdad, Baghdad, Iraq.}%
\thanks{This work has received funding from the Swedish Knowledge Foundation (KKS) project ``Semantic Robots''
}%
}

\begin{document}

\maketitle
\thispagestyle{empty}
\pagestyle{empty}

\begin{abstract}

This paper presents a new detector for
filtering noise from true detections in radar data, which improves the state of the art in radar odometry.
Scanning \ac{FMCW} radars can be useful for localisation and mapping in low visibility, but return a lot of noise compared to (more commonly used) 
lidar,
which makes the detection task more challenging. 
Our \ac{BFAR} detector  is different from the classical \ac{CFAR} detector in that it applies an affine transformation on the estimated noise level
after which the parameters that minimize the estimation error
can be learned.
\ac{BFAR} is 
an optimized combination between \ac{CFAR} and fixed-level thresholding. 
Only a single parameter needs to be learned from a training dataset.
We apply \ac{BFAR} to the use case of radar odometry, and adapt a state-of-the-art odometry pipeline (\acs{CFEAR}), 
replacing its original conservative filtering with \ac{BFAR}. 
In this way we reduce the state-of-the-art translation/rotation odometry errors
from 1.76\%/0.5$^\circ$/100~m to 1.55\%/0.46$^\circ$/100~m; an improvement of 12.5\%.
\end{abstract}

\section{Introduction}\label{sec:intro}

\begin{figure}
 \centering
 \includegraphics[trim={0cm 0cm 11.5cm 0cm},clip,width=1.0\hsize]{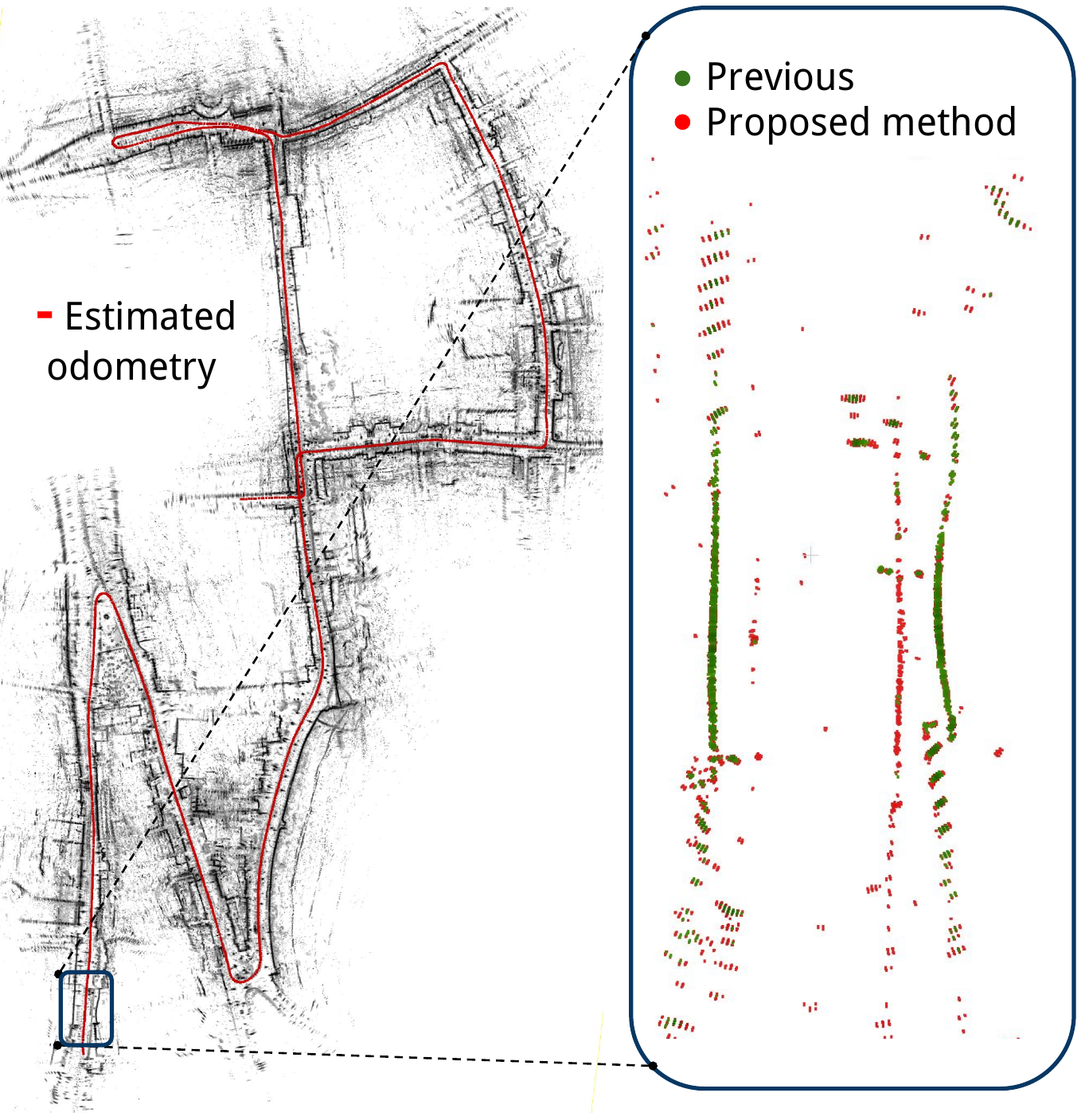}
 \caption{The Proposed radar detector BFAR yields high detection quality and can be used to improve radar odometry pipelines such as ``\ac{CFEAR}''.~\cite{adolfsson2021cfear}\label{fig:front_page}}
\end{figure}

There is a great need for enabling mobile robots and self-driving cars to robustly operate in all seasons, weather and visibility conditions. While these are challenging conditions for vision and lidar, \acp{FMCWr} are not affected by smoke, dust, rain or fog, and can detect obstacles regardless of visibility settings. Recently, radars, and especially spinning \ac{FMCW} radars have become compact, accurate and gained popularity and demonstrated more resilient localisation and mapping~\cite{hong2020radarslam}. This makes them suitable for applications in harsh environments; e.g., underground mines~\cite{brooker2007seeing} and fire fighting~\cite{mielle-2019-comparative} tasks. Unfortunately, radar suffer from high level of noise and clutter from multipath reflections. Previous filters have required tuning of several parameters, been sensitive to changes in environment~\cite{barnes_masking_2020,burnett2021radar}, required an extensive amount of training data~\cite{barnes_masking_2020,barnes_under_2020} or provided too few detections~\cite{adolfsson2021cfear,marck2013indoor}. For that reason, we present a novel target detection \ac{BFAR} that improve on existing noise filtering techniques in order to detect only \emph{useful} radar features. We demonstrate our filter on the previously existing pipeline for radar odometry ~\ac{CFEAR}, depicted in Fig.~\ref{fig:front_page}), which improves state-on-the-art from 1.76\% to 1.54\% translation error on the Oxford Radar RobotCar dataset (an improvement of 12.5\%).

The paper is structured as follows. 
\Section~\ref{sec:problem} gives a more in-depth description of the common problems with radar detection, and gives an intuition for our solution. 
\Section~\ref{sec:relatedWork} covers related work. 
Then in \Section~\ref{sec:problem-formulation} we present \ac{BFAR} in detail, a parameter estimation procedure and the adopted odometry pipeline \ac{CFEAR}.
Finally, results and conclusions are discussed in \Sections~\ref{sec:Field Results} and \Section~\ref{sec:Conclusion}.

\section{Problem formulation}\label{sec:problem}
Target detection from radar signals faces several challenges: the unstable and highly variable noise floor, speckle noise, ghost targets (false positive detection due to multi-path reflections), receiver saturation, and finally that highly reflective targets usually appear in several range-cells instead of only one, which reduces accuracy. We propose a new detector for radar signals 
and show how it
improves radar-only odometry estimation. Even though 
we apply this detector specifically for \ac{FMCWr} in this paper,
it can be extended to many other applications that have similar noise issues. 

\begin{figure}
	\centering
	\includegraphics[]{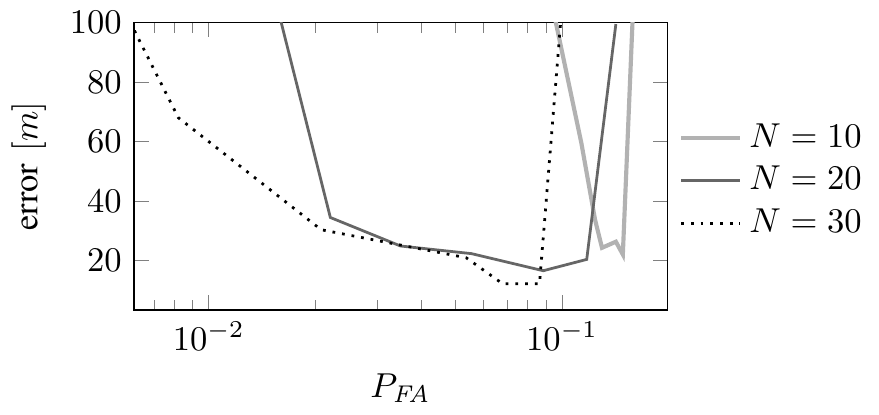}
	\caption{The sensitivity of odometry estimation absolute trajectory error vs \ac{PFA} values of the \ac{CACFAR} detector. The plots corresponding to an example from the Oxford RobotCar dataset~\cite{RobotCarDatasetIJRR}. The parameter $N$ is half the sliding window length when computing the cell averaging. }
	\label{fig:cfar_problem}
\end{figure}

There is an agreement among researchers on radar odometry that the classical \ac{CFAR} detection methods 
are not directly suitable for this application. For example, the \ac{CACFAR} detector, a common variant of \ac{CFAR} that computes a threshold for each cell that is proportional to the average of surrounding cells in a sliding window analysis, is not very useful with the odometry estimation pipeline in~\cite{adolfsson2021cfear}. It requires specifying precisely the values of \ac{PFA}, within a limited range, in order to have the algorithm working properly. Larger values of \ac{PFA} will return many detections from noise which makes odometry estimation much more challenging, while smaller values will return very few detections which also will cause odometry estimation to fail. \Figure~\ref{fig:cfar_problem} presents the odometry estimation error against \ac{PFA} values for \ac{CACFAR}. 
It shows that the error rapidly increases outside a certain \ac{PFA} range. Unfortunately, this range is not known, and it is hard to predict. Therefore, 
using point clouds filtered with \ac{CFAR} for radar odometry tends to fail,
unless one has a good and data-specific idea about selecting a suitable value 
(as also demonstrated in \Section\ref{subs:Estimation performance compared to}).
Our proposed \ac{BFAR} method 
makes the selection of \ac{PFA} much more flexible. 

\begin{figure}
	\centering	
	\includegraphics[]{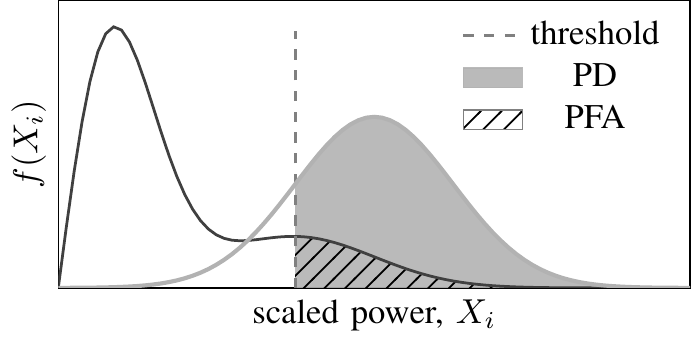}
	\caption{Graphical representation of the \ac{PFA} and \ac{PD}, the shaded areas, for a single sample $X_i$. The black density is the \ac{PDF} of $X_i$ when there is only noise, while the gray density is the \ac{PDF} of $X_i$ when there is a target.}
	\label{fig:pfa-pdet}
\end{figure}

The design of a detector is usually started from specifying the \ac{PFA}. The optimum detector is the one that maximizes the \ac{PD} for a given \ac{PFA}. (See \Figure~\ref{fig:pfa-pdet} for graphical illustration). In a stationary noise environment, a fixed-threshold detector can achieve the optimum detector performance if the noise power is known. In practice, the noise power is not known, therefore a statistical estimator is used to estimate its value. Additionally, the noise is usually mixed with clutter (reflections from objects other than the target: buildings, trees, ground surface, etc.) and therefore is non-stationary. For both reasons, the fixed threshold is replaced by an adaptive detector, usually a \ac{CFAR} variant, to have a controlled \ac{FAR}.
In the literature, there are many detection algorithms for radar signals that are based on the \ac{CFAR} detector strategy. The basic idea of \ac{CFAR} is to set a different threshold $T$ for each \ac{CUT}, using a constant multiplied by the estimated noise level at that cell. The constant is computed from a specified \ac{FAR} and the estimated noise level $Z$ is found using some statistical analysis on the $N$ neighboring cells.
\begin{equation}
	T
	=
	\text{constant} \times Z
\end{equation}
This scheme has the advantage of changing the threshold for each cell according to the estimated noise level but also has a clear drawback; since $Z$ is a random variable with variance, a large constant multiplier results in an amplified threshold variance by, the square of the constant. This scheme fails to achieve the specified constant \ac{FAR} when we have a bad estimator of the noise level from neighboring cells and the actual \ac{FAR} could be larger or smaller.
 
\begin{figure}
	\centering
	\includegraphics[]{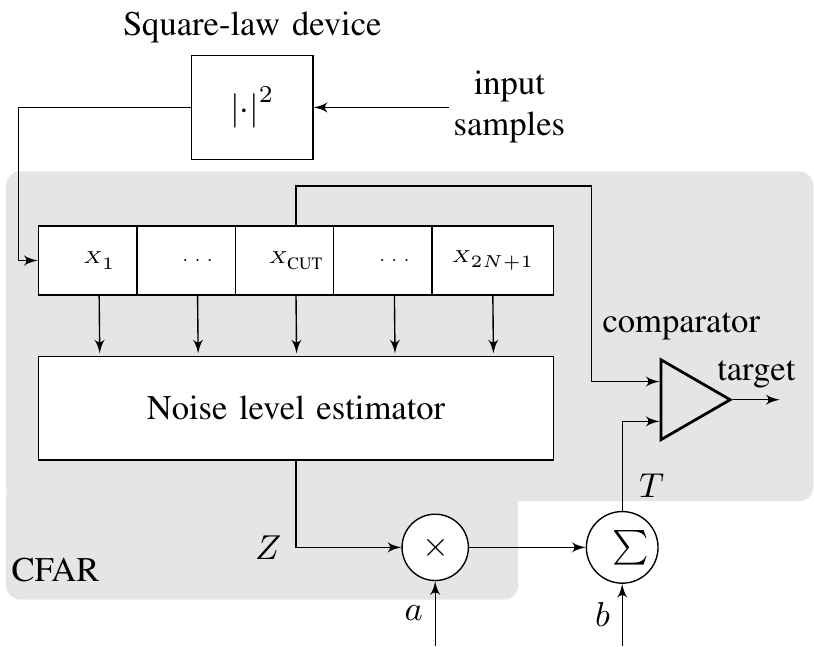}
	\caption{The structure of \ac{BFAR} detector. It is at a glance similar to \ac{CFAR} but fundamentally different as it applies linear transformation on the estimated noise level. Notice that, in this manuscript, we always use four guard-samples two on each side of $X_{\text{CUT}}$ which  are not shown in this figure to perceive the simplicity.}
	\label{fig:BFAR_detector}
\end{figure}
In this paper, we present a new detection algorithm that is based on a different strategy than \ac{CFAR}. It can handle the variable noise floor in a better way than the standard \ac{CFAR} and does so 
by selecting a threshold from an affine transformation of the estimated noise level $Z$, as explained in \Figure~\ref{fig:BFAR_detector}. The transformation requires two parameters: \emph{scale} $a$ and \emph{offset} $b$. The scale is selected to ensure that the threshold variance does not increase dramatically, while the offset parameter is selected to control \ac{PFA}. In this setup, both \ac{PD} and \ac{PFA} will depend on the unknown noise power, which might seem to be undesirable. However, their values will be controlled in such a way that \ac{PFA} and \ac{PD} will decrease as the noise power is decreased, and they increase with increased noise but will not exceed a specified upper bound. Most importantly, this control is done automatically as will be shown in \Section~\ref{sec:problem-formulation} when deriving the \ac{PFA} equation. 
In other words,
instead of maintaining \ac{PFA} constant but with large variance as in \ac{CFAR}, \ac{BFAR} fixes the \ac{PFAB} and sets \ac{PFA} with lower variance compared to \ac{CFAR}.

The threshold in \ac{BFAR} for each cell is then
\begin{equation}
	T
	=
	a \times Z + b.
\end{equation}
The affine transformation will not give a constant \ac{FAR} but will give \ac{FAR} that is bounded from above -- hence the name \ac{BFAR}. Learning the best values for the parameters can be done using grid search on a training dataset, (as demonstrated in \Section~\ref{ssec:Parameters procedure}). We assume here the values of $a$ and $b$ are fixed and not changing during the estimation process.

Notice that in \Figure~\ref{fig:BFAR_detector} the \emph{noise level estimator} block is generic, therefore it is possible to apply the \ac{BFAR} structure on most \ac{CFAR} variants by plugging-in the corresponding noise estimation block without further modifications.

\begin{figure}
    \centering
    \includegraphics[]{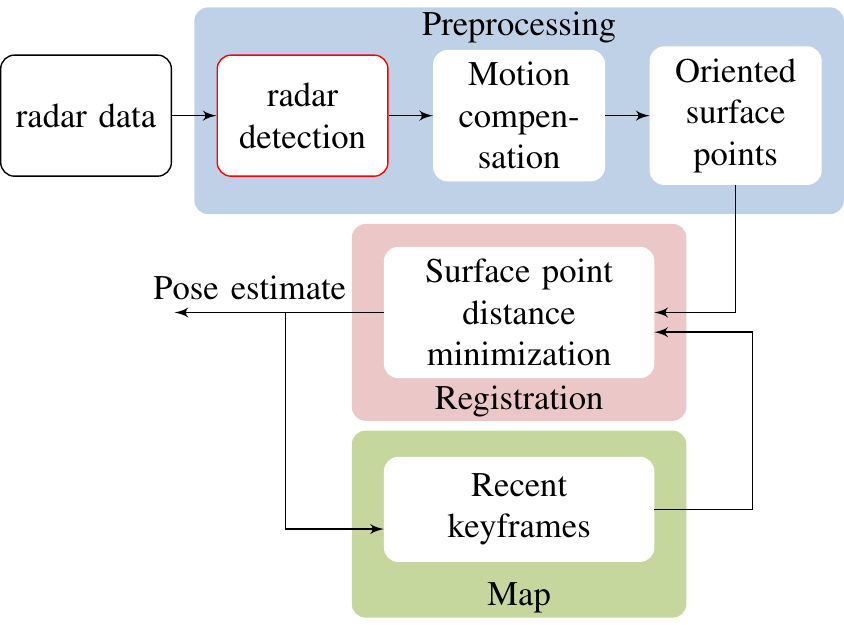}
    \caption{Overview of \ac{CFEAR} radar odometry, used for evaluating \ac{BFAR} radar detection.
    In the previously published version of CFEAR~\cite{adolfsson2021cfear}, the \emph{radar detection} block uses a $k$-strongest detector. 
    We replace this block with \ac{BFAR} and show how it leads to a new state of the art in  radar odometry.
    }
    \label{fig:cfear_radarodometry}
\end{figure}
To evaluate and examine the performance of \ac{BFAR}, we include it in the recent radar odometry pipeline CFEAR~\cite{adolfsson2021cfear}, 
depicted in Fig.~\ref{fig:cfear_radarodometry}. CFEAR is an efficient and accurate pipeline for incremental pose estimation that operates on $360^\circ$ spinning radar data, producing state-of-the-art accuracy in terms of low odometry drift.

CFEAR extracts features in a
two-step approach that first applies a conservative filter in polar space, keeping the $k$ strongest returns for each azimuth angle,  and then computes a sparse set of oriented surface points in Cartesian space. The sparse set of surface points is registered jointly to a history of key frames in order to estimate the pose and velocity of the radar. The estimated velocity is used to perform motion compensation.

In the original publication~\cite{adolfsson2021cfear}, CFEAR radar odometry 
was optimized to be fast and efficient, using a conservative $k$-strongest filter (with $k=12$ detections per azimuth) that typically removes secondary landmarks, even though these could be used for increased accuracy. For that reason, the conservative filter can be replaced with our proposed filter, which produces improved detections compared to the conservative filter, and can additionally aid odometry estimation. 

We use odometry quality as an indirect measure of detection performance, assuming that a good detector produces consistent and stable landmarks and are suitable for odometry. However, our method should be applicable for other perception tasks.

In \Section~\ref{sec:Field Results} we show quantitative results on the Oxford Radar RobotCar dataset and demonstrate that \ac{CFEAR} radar odometry with \ac{BFAR} detection instead of $k$-strongest produces new state-of-the-art accuracy for radar-only odometry.

\section{Related work}
\label{sec:relatedWork}

\subsection{radar only odometry estimation}
Previous methods for radar odometry overcome radar noise by learning to extract key points~\cite{burnett2021radar,barnes_under_2020}, remove noise~\cite{aldera2019fast,barnes_masking_2020,burnett2021radar,aldera2019fast} -- or by applying learning-free approaches that analyze intensity of reflections~\cite{cen2018precise,cen2019radar,adolfsson2021cfear,hong2020radarslam,kung2021normal,hong2020radarslam}.

Filter free approaches for odometry were investigated by Hong~\cite{hong2020radarslam} et al., who extracted and matched SURF features; and Park~\cite{park2020pharao} et al., who proposed a method for dense matching of raw radar images. However, these methods have demonstrated relatively high odometry errors compared to the current state of the art.
A similar dense matching technique was proposed by Barnes et al., which additionally uses self-supervision from ground truth poses for learning to mask out noise. Unfortunately, the odometry quality is reduced when operating in environments that was not in the training set, despite using a large amount of training data. Kung et al. applied a fixed intensity threshold to discard weak potentially spurious reflections~\cite{kung2021normal}. 
A more refined, though still ad-hoc, method, was successfully used by Adolfsson et al.~\cite{adolfsson2021cfear}, which  
additionally bounds the number of returned reflections to improve efficiency.
Cen et al.~\cite{cen2018precise,cen2019radar} investigated approaches for extracting features by analyzing image intensity and gradients.

Until today, the top performing method for radar odometry is \emph{\ac{CFEAR}}, despite that the original publication integrates a conservative filter designed to omit non-primary, but potentially important landmarks~\cite{adolfsson2021cfear}. In the present work, we replace their conservative filter with our proposed \ac{BFAR} method for higher quality detections, and accordingly, improve odometry estimation. 

\subsection{fixed-level detector}
Some authors have used simple fixed-level thresholding instead of an adaptive method like CFAR~\cite{hong2020radarslam,kung2021normal,mielle-2019-comparative}, however, selecting a suitable threshold value is not trivial. For example, Hong~\cite{hong2020radarslam} extracted only peaks that are greater than a standard deviation of mean intensity per azimuth. Kung et al.~\cite{kung2021normal} suggest keeping all points exceeding a noise threshold, which needs to be learned offline. Adolfsson et al.~\cite{adolfsson2021cfear} showed that this can have a negative impact on efficiency and may have negative impact on map quality.
Other researchers suggested conservative filtering~\cite{marck2013indoor,adolfsson2021cfear}. The most restricted one, used by Marck et al.~\cite{marck2013indoor}, keeps only the strongest reflection per azimuth, which will neglect a lot of potentially important information and possible good landmarks. Adolfsson et al. in \ac{CFEAR}~\cite{adolfsson2021cfear} relaxed this a little bit to keep $k$ strongest returns per azimuth, using values of $k=6\sim12$. They presented very good results in terms of translation and rotation errors on the Oxford public robotcar dataset~\cite{RobotCarDatasetIJRR}. For this reason, and also to make the results directly comparable, we will use the same odometry estimation pipeline presented in~\cite{adolfsson2021cfear} but replacing the $k$-strongest filter with our proposed detector.


\subsection{\ac{CFAR} detector}
The principle of \ac{CFAR} was first described in 1968 by Finn and Johnson \cite{finn1968adaptive}, they presented what is now known as \ac
{CACFAR}. Since then, \ac{CFAR} has been extensively studied in the literature and many variants, more than 25~\cite{machado2017evaluation}, have been proposed. 
We will give a brief review on \ac{CFAR} since it is the main part of the proposed detector.
The main challenges facing \ac{CFAR}~\cite{vivet2013localization} detectors are the following.
\begin{itemize}
	\item Mutual-target-masking, preventing the detection due to threshold increasing when a target falls within the reference cells. \ac{SOCACFAR}, \ac{TMCFAR}~\cite{gandhi1988analysis}, \ac{CSCFAR} and \ac{OSCFAR}~\cite{levanon1988detection} solve the mutual-target-masking problem \cite{rickard1977adaptive}.
	\item Clutter boundaries,  the abrupt change in the interference power -- solved using \ac{GOCACFAR}.
	\item Multiple-targets in the neighboring cells.
	\item Non-homogeneous noise or clutter. \cite{gandhi1988analysis,himonas1992automatic,farrouki2005automatic,shin2020robust}. 
\end{itemize}
The last two are still an active research directions and many researchers contribute continuously. One reason behind that could be the new development in sensors technology and the new challenging noise therein.



The performance of the \ac{CACFAR} detector was presented in \cite{rickard1977adaptive} for one or more interfering targets presented in the noise-level estimating cells. 
They also presented 
\emph{censored \ac{CFAR}} 
to maintain acceptable performance in the presence of interfering targets. 
An analysis of the performance of \ac{CACFAR}, \ac{GOCACFAR}, \ac{SOCACFAR}, \ac{OSCFAR}, and \ac{TMCFAR} in homogeneous and nonhomogeneous backgrounds was presented in \cite{gandhi1988analysis}. They computed the average detection threshold for each scheme to compare detection performance. To obtain closed-form expressions, they used an exponential noise model for clear and clutter backgrounds.

Automatic censored \ac{CFAR} detection for nonhomogeneous environments was first presented  in \cite{himonas1992automatic}, and a variability index to distinguish between homogeneous and nonhomogeneous noise in~\cite{farrouki2005automatic}. 
The \ac{OSCFAR} analyzed in \cite{levanon1988detection} for \ac{CFAR} loss, \ac{PFA}, and \ac{PD}. The performance analysis of \ac{CMLDCFAR} presented in~\cite{ritcey1986performance} and the \ac{TMCFAR} detector, a generalization of \ac{OSCFAR} and \ac{CMLDCFAR} detector presented in~\cite{gandhi1988analysis}. 

Recently, the \ac{OSCFAR} and its derivatives are the most commonly used for detection. The developments mainly focus on how to distinguish between cells from homogeneous noise and others from nonhomogeneous noise to decide which cell to keep and which to remove from the estimation process. Automatic censoring based on ordered data difference was proposed in \cite{jiang2016automatic}. A robust control scheme for adjusting detection threshold of \ac{CFAR} presented in~\cite{shin2020robust}. 
The cell-averaging clutter-map (CA-CM) with variability indexes suggested by~\cite{yang2020anti}. 

The structure of \ac{BFAR} presented in \Figure~\ref{fig:BFAR_detector} makes it possible to use most of the \ac{CFAR} variants above, however, we will consider only \ac{CACFAR} for simplicity and a proof of concept.

\section{\acf{BFAR}}
\label{sec:problem-formulation}

\subsection{Mathematical derivation}
In this section we will give the derivation of \ac{PFA} for our proposed \ac{BFAR} detector. We will assume the cell-averaging noise estimation in \Figure~\ref{fig:BFAR_detector} for simplicity and comparison purposes with \ac{CACFAR}.

We will use the following assumptions:
\begin{itemize}
	\item Assuming independent and Gaussian noise, then the output of square-low detector will be exponentially distributed and independent, with \ac{PDF}
	\begin{equation}
		f(X_i)
		=
		\frac{1}{2\lambda}\text{exp}\left(\frac{-X_i}{2\lambda}\right), \qquad x\ge 0
	\end{equation}
	and
	\begin{equation}
		\lambda
		=
		\begin{cases}
			\mu&,\qquad\text{under H}_0\\
			\mu(S+1)&,\qquad\text{under H}_1
		\end{cases} 
	\end{equation}
	where
	\begin{itemize}
		\item H$_0$ is the null hypothesis of no target, so $X_i$ contains only noise,
		\item H$_1$ is the alternative hypothesis of presence of a target,
		\item $S$ is the average \ac{SNR} of a target,
		\item $\mu$ is the total background clutter-plus-thermal noise power, its value is not fixed and is changing from cell-to-cell according to unknown but continuous and smooth function (no fast changes are expected).
	\end{itemize}

	\item The detector performance is determined by the average detection and false alarm probabilities. We will follow the same procedure as in~\cite{gandhi1988analysis}, the \ac{PFA} is, (see also \Figure~\ref{fig:pfa-pdet})
	\begin{equation}
		\text{\ac{PFA}}
		=
		\ExpectationOfGiven{\ProbabilityDensityOfGiven{X_{\text{CUT}}>aZ+b}{\text{H}_0}}{Z}
	\end{equation}
\begin{eqnarray}
		\text{\ac{PFA}}
		&=&
		\ExpectationOf{\int_{aZ+b}^{\infty}
			\frac{1}{2\mu}\text{exp}\left(\frac{-y}{2\mu}\right)dy
		}\\
		&=&
		\ExpectationOf{
			\text{exp}\left(-\frac{aZ+b}{2\mu}\right)
		}\\
		&=&
		\ExpectationOf{
		\text{exp}\left(-\frac{aZ}{2\mu}\right)
		} 	\ExpectationOf{
		\text{exp}\left(\frac{-b}{2\mu}\right)
		}\\
		&=&
		M_Z\left(\frac{a}{2\mu}\right)\text{exp}\left(\frac{-b}{2\mu}\right)
		\label{equ:mgf}
\end{eqnarray}
	where $M_Z(\cdot)$ denotes the \ac{MGF} of the random variable $Z$.
	\item For a \emph{cell-averaging} processor operating on $2N$ samples,\footnote{Here we are considering the \ac{CACFAR} for simplicity only. Please note that any other processors like \ac{OSCFAR}, \ac{TMCFAR}, etc. can be used in the same way.} the random variable $Z$ will be
	\begin{equation}
		Z
		\DefinedAs
		\sum_{i=1}^{2N} X_i
	\end{equation}
where $X_i$ are the samples of the neighboring cells. From~\cite{gandhi1988analysis}, the distribution of $Z$ will be
	\begin{equation}
	Z \sim \GammaDistribution{2N}{2\mu},
	\end{equation}
	 and the corresponding \ac{MGF} is $$M_y(u) = \left(1+2\mu u\right)^{-2N}.$$Substituting this in \Equation~\eqref{equ:mgf} above, we obtain
	 \begin{equation}
	 	\text{\ac{PFA}}
	 	=
	 	\left(1+a\right)^{-2N} \text{exp}\left(\frac{-b}{2\mu}\right).
	 	\label{equ:pfa}
	 \end{equation}
\end{itemize}
Setting the parameter $b = 0$, \Equation~\eqref{equ:pfa} will be reduced to
	 \begin{equation}
	\text{\ac{PFA}}
	=
	\left(1+a\right)^{-2N} \qquad \text{for}~b=0
	\label{equ:pfa_b_0}
\end{equation}
which is exactly the \ac{PFA} equation for \ac{CACFAR}, (the expression and derivation in~\cite{gandhi1988analysis}). Setting $a=0$ will give
	 \begin{equation}
	\text{\ac{PFA}}
	=
	\text{exp}\left(\frac{-b}{2\mu}\right) \qquad \text{for}~a=0
	\label{equ:pfa_a_0}
\end{equation}
which is the \ac{PFA} for a fixed-level detector. So \ac{BFAR} with \emph{cell-averaging} can be seen as a generalization for the classical \ac{CACFAR} detector. The smaller values of $\mu$ compared to $b$ correspond to a smaller exponential term in \eqref{equ:pfa} which makes the value of \ac{PFA} even smaller. On the other hand, very large values of $\mu$ compared to $b$ make the exponential term closer to one and the \ac{PFA} simplifies to
\begin{equation}
	\text{\ac{PFA}}
		\approxeq
	\text{\ac{PFA}}_{ub}
	\DefinedAs
	\left(1+a\right)^{-2N}
	\label{equ:pfa_approx}
\end{equation}
 which is the upper-bound for its value. Then we rewrite \eqref{equ:pfa} as
	 \begin{equation}
	\text{\ac{PFA}}
	=
	\text{\ac{PFA}}_{ub}~\text{exp}\left(\frac{-b}{2\mu}\right).
	\label{equ:pfa_new}
\end{equation}
When the parameter $a = 0$, both \ac{PFA} and \ac{PD} will be directly related to the background noise level $\mu$.

In a similar way we derive an expression for \ac{PD}:
\begin{eqnarray}
	\text{\ac{PD}}
	&=&
	\ExpectationOfGiven{\ProbabilityDensityOfGiven{X_{\text{CUT}}>aZ+b}{\text{H}_1}}{Z}\\
	&=&
	 \left(1+\frac{a}{1+S}\right)^{-2N}\text{exp}\left(\frac{-b}{2\mu\left(1+S\right)}\right).
	\label{equ:p_of_detection}
\end{eqnarray}
The exponential term in~\eqref{equ:p_of_detection} will indeed reduces \ac{PD} compared to \ac{CACFAR}.

\subsection{Parameter learning}
\label{ssec:Parameters procedure}

It is clear from \eqref{equ:pfa} that we need to tune $N$ and the two parameters $a$ and $b$ for the detector, which is not straightforward especially because the value of $\mu$ is not known. Therefore we propose the following procedure:
\begin{itemize}
	\item First, with a specified value of $N$ and \ac{PFAB}, the parameter $a$ can be computed from the \ac{PFAB} using \eqref{equ:pfa_approx},
	\item Second, the parameter $b$ can be learned from a use-case relevant data set. For the present paper, we are mainly interested in radar odometry, and use a training data set with ground-truth pose information (from GNSS) and learn $b$ using a suitable grid search method
	\item Finally, use the obtained parameter values in our detection algorithm for radar data. 
\end{itemize}

\section{Experimental validation}
\label{sec:Field Results}
\subsection{Methodology}
We investigated the performance of the proposed detection algorithm on odometry estimation using the Oxford Radar RobotCar dataset \cite{RadarRobotCarDatasetArXiv,RobotCarDatasetIJRR}. It has both radar raw data and ground-truth information for the odometry. The evaluation is performed using the standard odometry benchmark from KITTI~\cite{Geiger2012CVPR} to compute the following performance metrics:
\begin{itemize}
	\item Average sub-sequence translational and rotational errors, by evaluating possible sub-sequences of length (100, 200, $\ldots$,800) meters.
	\item \ac{ATE}: the \ac{RMSE} between estimated poses and corresponding ground truth.
\end{itemize}

\begin{figure}
	\centering
	\includegraphics[]{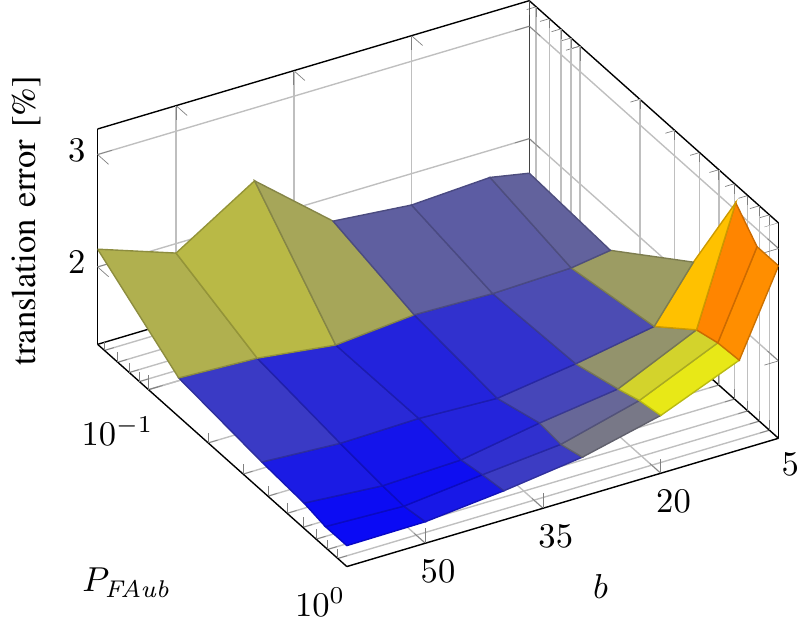}
	\caption{Translation error surface with \ac{PFAB} and $b$ for $N=20$. }
	\label{fig:a_b_transl}
\end{figure}

\label{ssec:learning the parameter b}
To learn the offset parameter $b$,
we select one sequence from the dataset.
With $N=20$ and parameter grid $a = [0, 0.1, 0.25, 0.5, 1, 2, 3]$, $b = [5, 10, 15, 20, 30, 40, 50, 60]$, we record the odometry metrics to find the best value for the parameters. As stated above, since $N$ is fixed the values of \ac{PFAB} can be computed directly from $a$ to be $[1, 0.0115, 0.0003, 9.5\times10^{-07}, 2.9\times10^{-10},9.1\times10^{-13}]$. The smallest translation error, rotation error and \ac{ATE} achieved during this grid search corresponds to $a=1$ or $\text{\ac{PFAB}}=9.5\times10^{-07}$ and $b = 20$ in the training dataset. See grid plot in \Figure~\ref{fig:a_b_transl}.

\begin{table*}
\begin{adjustbox}{width={\textwidth},totalheight={\textheight},keepaspectratio}%
	\centering
		
	\begin{tabular}{l|l|llllllll|ll}
	&&\multicolumn{8}{c}{Sequence}\\
	Method & resolution & 10-12-32 & 16-13-09 & 17-13-26 & 18-14-14 & 18-15-20 & 10-11-46 & 16-11-53 & 18-14-46 & mean & mean SCV\\
			\hline \hline
			\ac{CFEAR}-\ac{BFAR} (ours) & 0.043& \textbf{1.48/0.42} & \textbf{1.54/0.43} & \textbf{1.52/0.44} & \textbf{1.49/0.45} & \textbf{1.45/0.44} & \textbf{1.54/0.45} & \textbf{1.65/0.49} & \textbf{1.65/0.50} & \textbf{1.55/0.46} &  \textbf{1.55/0.46} \\
			\ac{CFEAR}-k12~\cite{adolfsson2021cfear} &0.043 & 1.64/0.48 & 1.86/0.52 & 1.66/0.48 & 1.71/0.49 & 1.75/0.51 & 1.65/0.48 & 1.99/0.53 & 1.79/0.5 & 1.76/0.50 &  1.76/0.50\\
            Cen2018~\cite{cen2018precise}& 0.175 & N/A & N/A & N/A & N/A & N/A & N/A & N/A & N/A & $3.72/0.95$& $3.63/0.96$\\
			Hong odometry~\cite{hong2020radarslam}&0.043 & 2.98/0.8 & 3.12/0.9 & 2.92/0.8 & 3.18/0.9 & 2.85/0.9 & 3.26/0.9 & 3.28/0.9 & 3.33/1 & 3.11/0.9 &  3.11/0.9\\
			Under the radar~\cite{barnes_under_2020}& 0.346& N/A & N/A & N/A & N/A & N/A & N/A & N/A & N/A & 2.0583/0.67 &  N/A\\
			Hero~\cite{burnett2021radar}& 0.2628& *1.77/0.62 & *1.75/0.59 & *2.04/0.73 & *1.83/0.61 & *2.20/0.77 & *2.14/0.71 & *2.01/0.61 & *1.97/0.65 & *1.96/0.66 &  N/A\\
			Barnes Dual Cart~\cite{barnes_masking_2020}& 0.043& N/A & N/A & N/A & N/A & N/A & N/A & N/A & N/A & 1.16/0.3 &  2.7848/0.0085\\
			Kung~\cite{kung2021normal}  & $0.125$ & N/A& N/A & N/A & N/A & N/A & N/A & N/A & N/A & 1.9584/0.6& 1.9584/0.6\\
			\hline
			SuMa (Lidar)~\cite{behley2018rss} & N/A& 1.1/0.3 & 1.2/0.4 & 1.1/0.3 & 0.9/0.1 & 1.0/0.2 & 1.1/0.3 & 0.9/0.3 & 1.0/0.1 & 1.16/0.3 & 1.16/0.3
			
		\end{tabular}
		\end{adjustbox}

	
	\caption{Evaluation on 8 sequences with different methods and sensor modalities on the Oxford radar RobotCar dataset~\cite{RadarRobotCarDatasetArXiv}. Results are given in (\% translation error / deg/100~m). ``*'' indicates that the results are not exactly comparable due to 
	lacking spatial cross validation.
	    }
	    \label{table:results2}
	\vspace{-0.3cm}
\end{table*}

\subsection{Odometry estimation performance compared to \ac{CACFAR}}
\label{subs:Estimation performance compared to}
\begin{figure}
	\centering
	\includegraphics[]{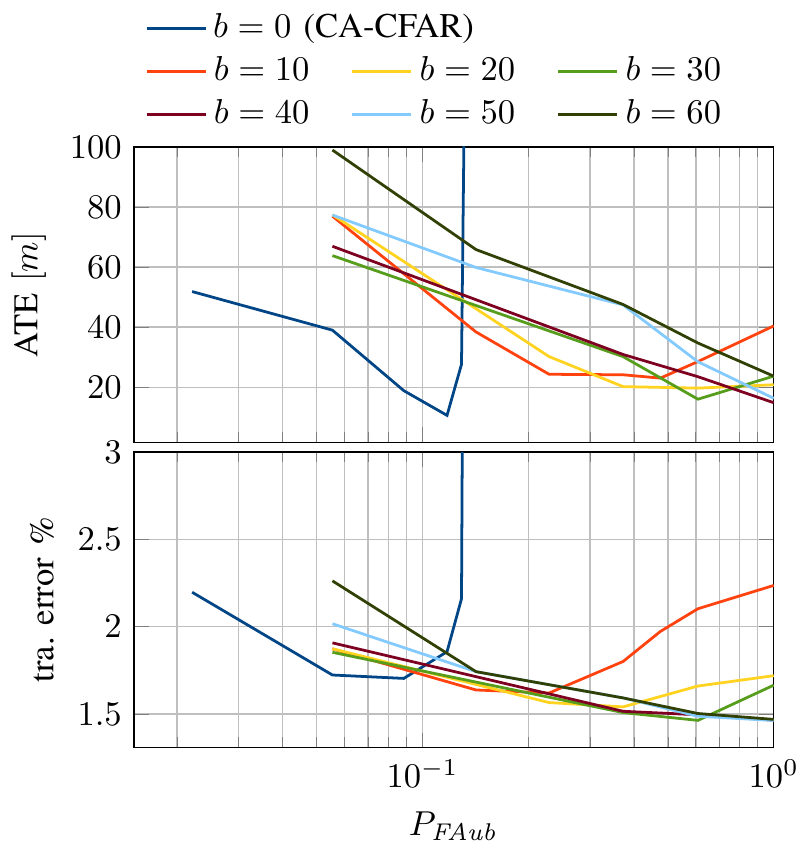}
	\caption{\ac{BFAR} estimation evaluation metrics vs the \ac{PFAB}.
	When $b=0$, \ac{BFAR} is the same as \ac{CACFAR}.
	}
	\label{fig:odometry_estimation_metrics}
\end{figure}

\Figure~\ref{fig:odometry_estimation_metrics} shows the plots of the different evaluation metrics vs \ac{PFAB} with parameter $b$. Notice that in the case of $b=0$ (blue curves)
, the detector will be \ac{CACFAR} and \ac{PFAB} is exactly equal to \ac{PFA} values. Comparing \ac{BFAR} with \ac{CFAR}, we can see that \ac{BFAR} works well even with $\ac{PFAB} =1$. Comparing this with the limited selection of \ac{PFA} in \ac{CFAR}, as presented in \Figure~\ref{fig:cfar_problem}, proves the flexibility in parameter selection and also the ability to automatically account for changes in background noise level. 
Notice that for $\ac{PFAB} =1$, it will be a fixed-level detector and the value of $b=40$ will give the smallest value for \ac{ATE}.

\subsection{Odometry estimation performance}
\label{subs:Odometry estimation performance}
We apply the learned and optimized parameters, $a=1$ and $b=20$, 
and insert \ac{BFAR} in the \ac{CFEAR} odometry pipeline for evaluation
on 
the remaining sequences
of the robotcar dataset. Except for the detector, we used the same parameters of \ac{CFEAR} as in the original publication~\cite{adolfsson2021cfear}.

The obtained odometry errors are presented in \Table~\ref{table:results2} and compared to seven recent radar odometry baselines, as well as one baseline using lidar odometry for reference. 
Compared to the state of the art, \ac{CFEAR} with \ac{BFAR} detection consistently has smaller error over all the trajectories.

\section{Conclusion}
\label{sec:Conclusion}
This paper focuses on radar detection for radar-only odometry estimation. 
We have proposed a new detector called \ac{BFAR} and showed that, compared to \ac{CACFAR}, its parameters can be easily trained such that it keeps relevant detections and removes radar noise.  
In particular, we have demonstrated the usefulness of \ac{BFAR} by incorporating it in the \ac{CFEAR} pipeline for radar odometry, which to date holds the lowest published odometry errors in the Oxford Radar RobotCar dataset.
Using the classical \ac{CACFAR} detector with \ac{CFEAR} will not lead to acceptable performance and the pipeline may fail. However, it was possible to achieve even smaller  estimation errors than the original \ac{CFEAR} when using our new \ac{BFAR} detector instead of the baseline $k$-strongest detector. 
%
Conceptually, \ac{BFAR} can be seen as an optimized combination between \ac{CACFAR} and fixed-level thresholding to minimize the odometry estimation error. 
It was possible to reduce the estimation translation/rotation errors in \ac{CFEAR} from 1.76\%/0.5$^\circ/100$m to 1.55\%/0.46$^\circ/100$m.

The structure of \ac{BFAR} makes it possible to use many variants of \ac{CFAR} to add their advantages, not necessarily the classical \ac{CACFAR} that we used here for simplicity and proof of concept. Finally, even though the \ac{BFAR} detector has been proposed specifically for spinning radar odometry estimation, it is expected to contribute in signal detection for other tasks and sensors with similar noise issues. 


\bibliographystyle	{../Bibliography/ieeetransactions_no_urls}		%
\bibliography		{../Bibliography/main}					%

\end{document}